%% file: tj-template-ap.tex
\documentclass{ieeetj}
\usepackage{lmodern}
\let\labelindent\relax
\usepackage[inline]{enumitem}
\usepackage{amsmath,amssymb,amsfonts}
\usepackage{bm}

\usepackage{hyperref}
\hypersetup{
    colorlinks=true,
    linkcolor=blue,
    filecolor=magenta,      
    urlcolor=cyan,
    }

\usepackage{cleveref}
\usepackage[normalem]{ulem}

\usepackage{tikz}

\usepackage{graphicx}
\usepackage{booktabs}
\usepackage{multirow}
\usepackage{rotating}
\usepackage{makecell}
\usepackage{color,colortbl}

\usepackage{enumitem}

\usepackage{textcomp}
\usepackage{xcolor}
\usepackage{physics}

\usepackage{algorithm}
\usepackage{algpseudocode} 

\usepackage{listings}

\usepackage{pgf}
\usepackage{pict2e}
\usepackage{wasysym}
\usepackage{stackengine}
\usepackage{comment}

\usepackage{cite}

\usepackage{cleveref}
\usepackage{amsmath,amssymb,amsfonts}
\usepackage{mathtools}
\usepackage{comment}
\usepackage{textcomp}
\usepackage{xcolor}
\usepackage{float}
\usepackage{soul}
\usepackage{cancel}
\usepackage{algorithm}
\usepackage{algpseudocode}
\usepackage{amsthm}
\usepackage{amsmath}
\usepackage{wrapfig}
\usepackage{color, colortbl}
\usepackage{stackengine}

\definecolor{LightCyan}{rgb}{0.4, 0.9, 0.8}

\usepackage{newtxtext,newtxmath}

\DeclarePairedDelimiter{\floor}{\lfloor}{\rfloor}

\newif\ifarxiv
\arxivtrue   

\ifarxiv
\makeatletter
\renewcommand{\p@subsection}{\thesection-}
\renewcommand{\p@subsubsection}{\thesection-\thesubsection}
\makeatother
\makeatletter

\def\@maketitle{\newpage
  \bgroup
    \vspace*{-10.5pt}%
    \vskip14.5pt{\titlefont{\textcolor{tmlcncolor}{\@title}}\par\vskip8.5pt}%
    {\authorfont\@author\par}\vskip4pt{\afffont\@affil}\par\@IEEEspecialpapernotice\vskip5pt%
    \ifx\@corresp\@empty\else{\corfont\@corresp\par}\vskip5.5pt\fi
    \ifx\@authornote\@empty\else{\afffont\@authornote\par}\vskip27.5pt\fi
    \centerline{\rule{36pc}{.5pt}}\par\vskip11.5pt
    \ifvoid\abstractbox\else{\reset@font\box\abstractbox}\fi\par\vskip11.5pt
    \ifvoid\keybox\else{\reset@font\box\keybox}\fi\par\vskip-32pt
    \centerline{\rule{36pc}{.5pt}}%
    \par\addvspace{33.5pt}\egroup
  \let\maketitle\relax}

\def\ps@plain{%
  \def\@oddhead{}\def\@evenhead{}%
  \def\@oddfoot{\hfil{\rffont\thepage}\hfil}%
  \let\@evenfoot\@oddfoot}

\def\ps@headings{%
  \def\@oddhead{}\def\@evenhead{}%
  \def\@oddfoot{\hfil{\rffont\thepage}\hfil}%
  \let\@evenfoot\@oddfoot}
\def\OJlogo{}
\def\seclogo{}

\makeatother
\fi

\begin{document}


\markboth{}{Lupu {et al.}: Semantic Segmentation and Closed-Loop Attitude Control}

\title{Learning in Orbit: Onboard Training and Closed-Loop Attitude Control with a Thermal Segmentation Model on a Spacecraft}

\renewcommand{\authorrefmark}[1]{\textsuperscript{#1}}

\author{Elena Sorina Lupu\authorrefmark{1,$\star$}, Hannah Grauer\authorrefmark{1,$\star$}, Phaedrus Leeds\authorrefmark{2,$\star$}, Nikhil Ranganathan\authorrefmark{1,$\star$}, Connor Lee\authorrefmark{1},
Sulekha Kishore\authorrefmark{1}, Cailyn Smith\authorrefmark{1}, Shon Cortes\authorrefmark{2},
Brian Hardy\authorrefmark{2}, Benjamen Bycroft\authorrefmark{2}, Darren W. Rowen\authorrefmark{2}, Soon-Jo Chung\authorrefmark{1} \\ }
\affil{California Institute of Technology, Pasadena, CA 91125, USA}
\affil{The Aerospace Corporation, 2310 E. El Segundo Blvd., El Segundo, CA 90245-4609, USA}
\authornote{This work was supported by The Aerospace Corporation. \\ $^\star$Co-first authors.}
\corresp{Corresponding author: Soon-Jo Chung (email: sjchung@caltech.edu).}



\begin{abstract}
Deep space missions increasingly require spacecraft capable of autonomously operating in environments with communication delays and bandwidth limitations.
A key enabler of such autonomy is the ability to run learning-based algorithms directly onboard. 
In this work, we present a flight deployment on a 6U CubeSat that demonstrates (1) onboard training and inference of a thermal segmentation model on an NVIDIA Jetson TX2 NX GPU in space and (2) closed-loop attitude control using the segmentation model on a real-time, thermal image feed to track the Moon.
In addition to this demonstration, we implement ROS 2 Foxy as a data transport middleware and integrate an open-source file system that provides strong data integrity guarantees and self-healing features. Over the course of approximately one year, we performed 12 experiments on the spacecraft, advancing the state of the art in autonomous spacecraft systems by showing how modern learning-based pipelines, commercially available hardware, and open-source software can operate reliably under the compute and environmental constraints of Low Earth Orbit (LEO).
\end{abstract}

\begin{IEEEkeywords}
Attitude control, Computer vision, Convolutional neural networks, Edge computing, Foundation models, Graphics processing units, Image segmentation, Infrared imaging, Kalman filters, Learning (artificial intelligence), Low Earth Orbit satellites, Robot operating systems, Target tracking.
\end{IEEEkeywords}


\maketitle

\section{INTRODUCTION}

\IEEEPARstart{O}{ne} of the goals of space exploration is to enable robotic systems to autonomously explore environments far from Earth~\cite{NASA_TA04_2015}.
Missions will increasingly rely on spacecraft capable of perceiving and learning from their environment and taking actions without a human-in-the-loop.
Such autonomy is essential for deep space exploration because communication delays, limited bandwidth, and mission complexity make reliance on ground operation impractical.

Achieving this goal requires spacecraft that can run learning algorithms for perception, estimation, planning, and control.
Advances in Artificial Intelligence (AI)~\cite{sutton2018reinforcement, lecun2015deep, SCHMIDHUBER201585} have already improved the capabilities of robotic systems on Earth, for example in quadcopters~\cite{neural_fly}, ground vehicles such as those developed by Waymo and Tesla, and walking robots~\cite{DBLP:journals/corr/abs-2010-09259, doi:10.1126/scirobotics.adi7566}.
These capabilities have become core components of autonomy stacks in terrestrial robotics, where perception models are tightly integrated with control systems. 


From a robotics perspective, space can be treated as another environment in which robotic systems must operate. 
Although physical conditions differ from those of Earth~\cite{lodders_planetary_1998}, perception, state estimation, planning, and control remain the same. The robotics community has made substantial progress in addressing these problems in terrestrial settings, and there is no fundamental reason why similar technologies cannot be deployed in space, provided that they are rigorously tested and their failure modes are characterized and mitigated at reasonable cost.

\begin{figure*}
    \centering
    \includegraphics[width=\linewidth]{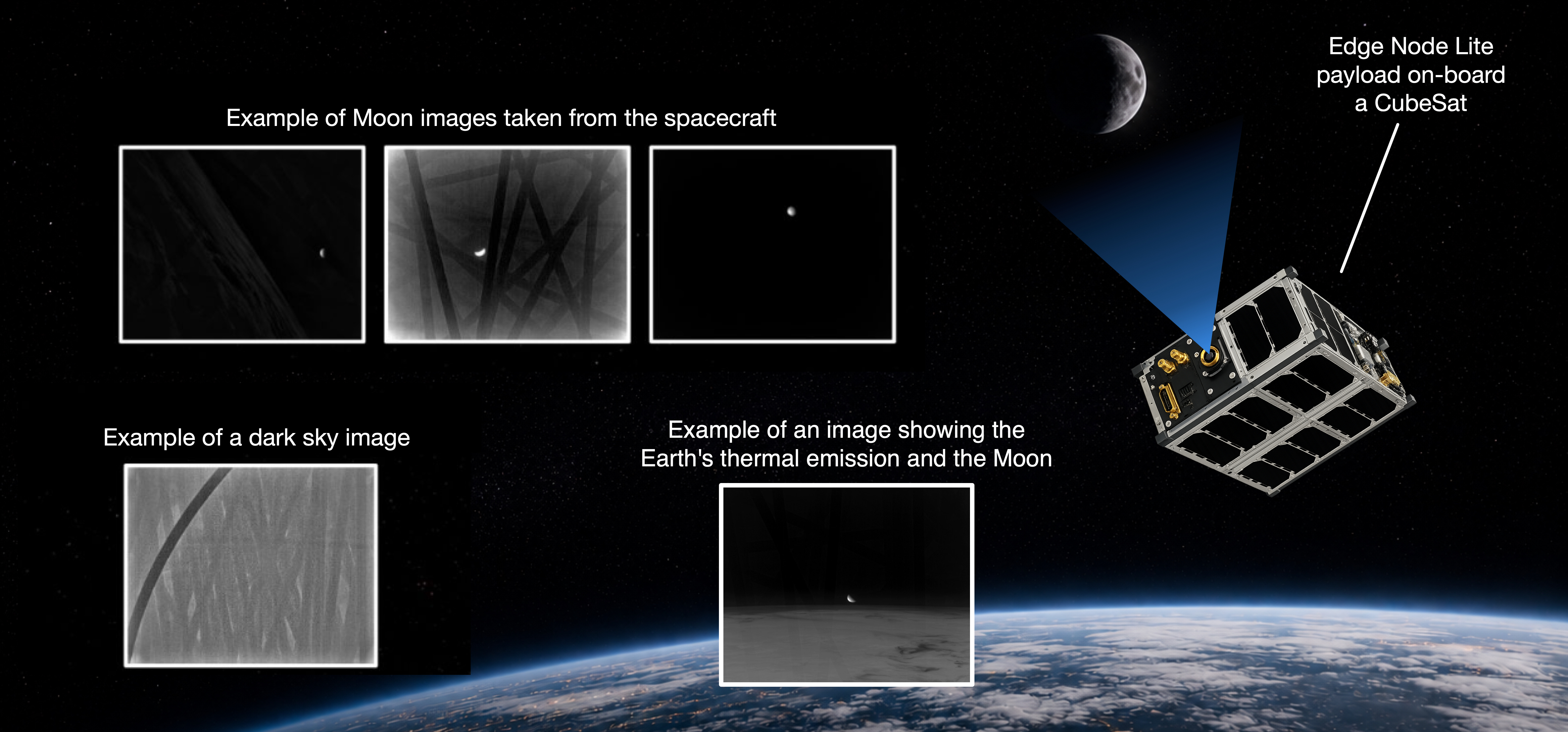}
    \caption{\textbf{The spacecraft, alongside images captured in flight.} 
    At the top, three Moon images captured by the Boson thermal camera are shown. On the left, one image taken of the dark sky exhibiting streak artifacts (see~\Cref{sec:camera_streaks}). In the middle, we show an example of a thermal image acquired in orbit, with the Earth's thermal emission and the Moon as a distinct thermal signature. The spacecraft is illustrated on the right.}
    \label{fig:architecture}
\end{figure*}

Spacecraft autonomy has historically relied on more classical algorithms due to the lack of available compute.
For example, the primary flight computer onboard the Perseverance rover (launched in 2020) is based on the RAD750 processor, a radiation-hardened CPU operating at  200~MHz and with 256~MB of RAM~\cite{nasa_perseverance_components}. 
Recently, Mars Helicopter used a Qualcomm Snapdragon 801 processor~\cite{mars_helicopter}, a commercial smartphone processor, showing that COTS compute can fly on Mars.
Other recent missions have demonstrated the feasibility of running machine learning (ML) models onboard spacecraft, but were focused on inference and training tasks on the CPU~\cite{demo_space, phi_sat}.

\subsection{Contributions}
In this work, we demonstrate that a spacecraft can learn and adapt in-flight, and use its learned perception models to act autonomously in orbit with minimal human supervision. Specifically, we present a flight demonstration of a closed-loop system in which a vision segmentation model is trained directly on a spacecraft in Low Earth Orbit (LEO) and used to command the spacecraft's attitude control system to point the spacecraft to a desired location, in this case the Moon. Our \textbf{primary contributions} are as follows:
\begin{itemize}[nosep]
    \item \textbf{onboard, vision-based model learning:} We successfully performed both training and inference on an NVIDIA Jetson TX2 NX's GPU on a CubeSat. Based on publicly available information, this marks the first time that machine learning models were trained on a GPU on a free-flying spacecraft\footnote{Machine learning models were trained in space before, but on low-power CPUs or VPUs, rather than GPUs~\cite{fastmodelinferencetraining}.}.
    \item \textbf{Closed-loop autonomy using learned models:} We integrated our segmentation model with the spacecraft’s Attitude Determination and Control System (ADCS) in order to autonomously command the spacecraft toward a target of interest (in this case, the Moon.)
\end{itemize}
We complete this study by providing \textbf{secondary contributions} documenting our findings of leveraging open-source software and commercially-available hardware to support deployments in LEO: 
\begin{itemize}[nosep]
    \item \textbf{ROS 2 in space:} We deployed the Foxy distribution of the Robot Operating System (ROS 2) as a middleware between the segmentation module and the bus avionics.
    To the best of our knowledge, ROS has not been demonstrated on a spacecraft free-flying in space before~\cite{Flckiger2019AstrobeeRS}.
    \item \textbf{File system:} We integrated an open-source file system (OpenZFS), which has strong guarantees around data corruption and repair as well as robustness to sudden power loss.
    \item \textbf{Thermal and CPU/GPU Usage:} We present measurements of the computer's thermal performance, as well as the usage of the CPU and GPU, during both inference and training.
\end{itemize}
This demonstration advances the state of the art in autonomous spacecraft systems by showing how modern learning-based pipelines, commercially available hardware, and open-source software can operate reliably under the compute and environmental constraints of LEO.


\section{RELATED WORK}

\subsection{Flown Missions with Learning Demonstrations}
Early demonstrations of AI\footnote{We understand AI from John McCarthy's definition ``Intelligence is the computational part of the ability to achieve goals in
the world'', machine learning (ML) as containing three forms of learning: supervised learning, unsupervised learning, and reinforcement learning, and deep learning (DL) as referring to using multi-layer artificial neural networks for function approximation.} in space focused on autonomous decision-making. One of the earliest examples is the Remote Agent (RA) experiment onboard Deep Space 1 (DS1), a comet flyby mission launched in 1998. This mission demonstrated closed-loop spacecraft control using symbolic reasoning and planning~\cite{MUSCETTOLA19985}.

One of the first uses of statistical machine learning (ML) onboard a spacecraft was on NASA’s Earth Observing-1 (EO-1) mission, launched in 2000~\cite{Schaffer2017CloudFA}. EO-1's Autonomous Sciencecraft Experiment incorporated multiple learning-based methods, including support vector machine (SVM) and hyperspectral unmixing techniques~\cite{6392936}. These methods were used to identify scientifically relevant observations.

Subsequent missions expanded the use of onboard learning by employing Deep Learning (DL). For example, CNES’s DEMETER mission (2004) employed a Time Delay Neural Network (TDNN)~\cite{cnes_2001}. This represented the first deployment of a neural network in space for scientific signal processing~\cite{Ekelund_2024}.
In the 2010s, the Intelligent Payload Experiment (IPEX) was built to validate technologies for onboard instrument processing and autonomous operations~\cite{ipex}.
IPEX demonstrated the use of support vector machine learning techniques, spectral unmixing, and random decision forest classification techniques. 
The IPEX random forest classifier was able to achieve high accuracy, despite being trained prior to launch using just four hand-labeled images from a high-altitude balloon test~\cite{ipex}.

Recent missions have increasingly leveraged deep learning and more capable onboard hardware. ESA’s OPS-SAT (2019) served as a flexible in-orbit laboratory for onboard computing and DL, supporting convolutional neural network (CNN) inference among other capabilities~\cite{demo_space} and demonstrated first re-training of an
onboard AI model with live in-flight data. Similarly, Phi-Sat-1 (2020) validated real-time cloud detection using a CNN running on an Intel Movidius Myriad 2 vision processing unit (VPU), enabling selective downlink of useful Earth observation data~\cite{phi_sat}.

Several commercial and experimental platforms have since extended these capabilities. D-Orbit’s ION Satellite Carrier demonstrated onboard training of ML models for applications such as cloud detection and disaster response using an Intel Myriad X processor~\cite{2023_orbit_demo}. Ubotica’s CogniSAT-6 (2024) further expanded the scope of onboard pipelines, supporting multiple Earth observation tasks including cloud screening, water monitoring, and land-use classification on similar hardware~\cite{ubotica}.  Spiral Blue’s SE-Z (2021) and SE-1 (2023) CubeSats also contributed to this trend by enabling onboard inference and image processing using embedded AI accelerators such as the NVIDIA Jetson Nano and Xavier NX platforms.

More recently, there has been a shift toward high-performance onboard compute platforms capable of supporting more complex AI.
Planet Labs’ Pelican-2 satellite (2025) incorporates an NVIDIA Jetson platform for real-time onboard analytics, although it remains unclear whether training, in addition to inference, is performed in orbit~\cite{planetLabs_2024}. Later in 2025, after our on-orbit training experiments in August 2025, NVIDIA-backed Starcloud trained a small language model on an H100 GPU in orbit, effectively operating as an orbital data center~\cite{nvidia_starcloud_2025}. 
A summary timeline of these missions is presented in~\Cref{tab:satellites}.

\begin{table*} 
\caption{Flown missions with onboard machine learning or AI capabilities (based on publicly available information). Early missions without neural networks are included for historical context as pioneering AI-driven autonomy in space; certain claims are reported as stated in the literature.
}
\centering \label{tab:satellites} \begin{tabular}{p{3cm}|p{2.3cm}|p{1.8cm}|p{2.8cm}|p{5.5cm}} \textbf{Mission} & \textbf{Platform} & \textbf{Launch Date} & \textbf{Hardware} & \textbf{Capability} \\  \midrule[1.0pt] 
Remote Agent (RA) on Deep Space 1 (DS1) ~\cite{MUSCETTOLA19985} & Medium-sized interplanetary spacecraft & Oct. 1998 & RAD6000~\cite{remote_agent_claim_first_ai} & ``The first AI-based closed loop autonomous control system to take control of a spacecraft''~\cite{remote_agent_claim_first_ai} \\ \hline 

NASA's Earth Observing-1 (EO-1)~\cite{Schaffer2017CloudFA} & Medium-sized Earth-observation satellite & Nov. 2000 & Mongoose-V~\cite{10.1109/SMC-IT.2006.48} & First ML in space~\cite{eo_1_first_ml_in_space} (used support vector machine (SVM) and later hyperspectral unmixing~\cite{6392936}) \\ \hline CNES's DEMETER~\cite{cnes_2001} & SmallSat & June 2004 & Analog Devices  ADSP 21020 DSP & ``First onboard neural network dedicated to the scientific processing of wave data'', Time Delay Neural Network~\cite{Ekelund_2024} \\ \hline Intelligent Payload Experiment~\cite{ipex} & Cubesat & Dec. 2013 & Gumstix Earth Storm & ML random decision forests to classify images onboard and CV visual salience software to extract interesting regions for downlink in acquired imagery~\cite{thompson-altinok-bornstein-et-al-2015}\\ \hline OPS-SAT~\cite{demo_space} & CubeSat & Dec. 2019 & Altera Cyclone V SX SoC (FPGA) & CNN image classification, k-means, FDIR model training~\cite{demo_space}. ``First re-training of an onboard AI model with live in-flight data''\\ \hline Phi-Sat-1~\cite{phi_sat} & CubeSat & Sep. 2020 & Intel Movidius board with a Myriad II chip (VPU) &  Inference of a CNN in space~\cite{phi_sat} \\ \hline 
Spiral Blue's SE-Z~\cite{spiralBlue0_2021} & CubeSat & July 2021 & NVIDIA Jetson Nano & In space processing of satellite images\\ \hline D-Orbit's ION SCV004~\cite{fastmodelinferencetraining} & SmallSat & Jan. 2022 & Intel Myriad X processor & Training of a cloud-detection ML model \\ \hline 
Spiral Blue's SE-1~\cite{spiralBlue1_2024} & CubeSat & Jan. 2023 & NVIDIA Xavier NX & Inference of an ML model in space \\ \hline ESA's MANTIS~\cite{merce2024mantis} & CubeSat & Nov. 2023 & Intel Movidius Myriad 2 (VPU) & ML processing onboard the satellite\\ \hline Ubotica's CogniSAT-6~\cite{ubotica} & CubeSat & March 2024 & Intel Myriad X processor & Onboard ship detection using ML and  control (dynamic targeting).  \\ \hline 
Phi-Sat-2~\cite{phi_sat_2} & CubeSat & Aug. 2024 & Intel Movidius Myriad 2 (VPU) & ML to process data directly in orbit  \\ \hline
PROMETHEUS-1~\cite{andrade2025prometheus1} & PocketQube & Jan. 2025 & STM32 & First ``onboard ML inference classification on a PocketQube''~\cite{andrade2025prometheus1} \\ \hline
Planet Labs Pelican-2~\cite{planetLabs_2024} & SmallSat & Jan. 2025 & NVIDIA Jetson & Real-time data analytics on Earth observation data~\cite{planet_pelican_2025} \\ \hline 
\end{tabular} 
\end{table*}

\subsection{Closed-loop Control using Vision}

Vision-based closed-loop control has been extensively studied in terrestrial robotics, where perception outputs are directly integrated into feedback control loops.
An early approach for including visual information for control is visual servoing~\cite{espiau1992new}, a technique mainly used for robot manipulation.
Recently, vision-based learning has demonstrated the capability to control agents in simulated environments\cite{world_models}, as well as robots in real environments, with applications to ground robots~\cite{magic_vfm, doi:10.1126/scirobotics.abk2822, pmlr-v205-wu23c} and robot manipulation~\cite{JMLR:v17:15-522, Florence2019SelfSupervisedCI, visualforesight}.

In space applications, vision has primarily been used for navigation, pose estimation, rendezvous and proximity operations (RPO), and Earth Observations (EO) processing. For example, SpaceX's Dragon Capsule uses its relative navigation sensors, such as thermal and LiDAR for docking. 
Similarly, vision-based navigation has been used for planetary landing and hazard detection, where onboard image processing informs guidance decisions~\cite{https://doi.org/10.1002/rob.70150}.

\subsection{ROS 2 Middleware}

Modular middleware architectures have played a central role in enabling robotic systems by decoupling perception, estimation, planning, and control components.
The Robot Operating System (ROS)~\cite{quigley2009ros} has become the de facto standard middleware in terrestrial robotics, providing an Observer pattern-based communication framework, hardware abstraction, and a rich ecosystem of tools for rapid development and integration.
Its successor, ROS 2~\cite{macenski2022ros2}, extends these capabilities with improved real-time performance and distributed communication via DDS (Data Distribution Service).

Reusable software frameworks such as core Flight System (cFS) exist for spacecraft flight software.
However, there currently does not exist a reusable and accessible flight software framework for robotic space systems~\cite{spaceros}.
Recognizing this gap, recent efforts have explored adapting ROS~2 for space environments.
NASA's Space ROS initiative~\cite{spaceros} aims to extend ROS~2 with features required for flight certification.
Despite these developments, there have been no demonstrations of ROS~2 running on free-flying spacecraft in orbit.

\section{THE EDGE NODE MISSION}

\subsection{Edge Node Lite}\label{sec:edge_node_lite}

The Edge Node mission is a distributed, multi-small-satellite system operating as a free-flying, collaborative testbed for formation flight and RPO.
It functions as a multi-user, on-orbit facility where emerging technologies and algorithms are  validated in LEO~\cite{EdgeNode}.

The Edge Node Lite mission is a precursor to the full Edge Node mission, with its primary objective being risk reduction through the demonstration of advanced onboard image processing and ML algorithms, including the approach presented in this work. Specifically, it contains a payload consisting of a subset of sensing and onboard computing hardware designed to validate key technologies in orbit.

A central goal of Edge Node is to accelerate the adoption of compact commercial-off-the-shelf (COTS)
computing and sensing technologies not originally
designed for space applications.
This includes leveraging modern computing hardware and open-source software frameworks.
Edge Node Lite demonstrated several of these capabilities, including training ML algorithms, the deployment of ROS 2 in orbit, and the use of containerized software via Docker.


The Edge Node Lite payload specifically consists of an NVIDIA Jetson TX2 NX paired with a 1 TB NVMe storage drive (WD IX SN530), a radiation-tolerant ARM Cortex-M7 microcontroller, a high-speed data interface to the spacecraft LASER downlink communication system, and a serial interface to the spacecraft bus including the attitude determination and control system (ADCS).
The NVIDIA Jetson TX2 NX operates on Ubuntu 18.04 and uses ROS 2 Foxy as the middleware.

The payload includes a FLIR Boson+ long-wave infrared (LWIR) camera.
The sensor is based on an uncooled VOx microbolometer operating in the 8–14 $\mu$m wavelength range.
This camera is used for the segmentation experiments presented in~\Cref{sec:results}.
The LWIR camera was chosen because of its day/night invariance. Some examples of images taken by the LWIR camera are shown in~\Cref{fig:architecture}.

\subsection{6U CubeSat}

The spacecraft is a 6U CubeSat that hosts the Edge Node Lite payload described in~\Cref{sec:edge_node_lite}. 
In addition to the payload, the satellite is equipped with an ADCS and solar panels, but does not include propulsion. 

The satellite operates in a sun-synchronous, near-polar low Earth orbit. 
 The orbit has an altitude of approximately 500 km, and an orbital period of about 94 minutes. The beta angle for the spacecraft varied from about $-26.2^\circ$ to $-16.0^\circ$ over the course of approximately a year. 
Positioning the spacecraft in a sun-synchronous orbit has the advantage that the same location on Earth is observed at nearly the same local solar time on every pass, which means that the heating and cooling cycles are similar.
The spacecraft is in eclipse for about 30\% of its orbit time.
To help with heat dissipation, the Jetson TX2 NX was thermally sunk to the wall of the spacecraft.

Edge Node Lite runs a customized version of NVIDIA's Linux for Tegra version 32.7.3, which is based on Ubuntu 18.04. The customizations include numerous Docker images: (a) system resource usage and temperature monitoring which is published to ROS 2, (b) software for collecting image frames, setting camera parameters, and collecting camera telemetry, also through ROS 2, (c) numerous ML  packages such as OpenCV, TensorFlow, and PyTorch in anticipation of use by experiments uplinked after launch, (d) neural networks (discussed in this paper) for computer vision. Other modifications to the operating system include configuration of disk-backed swap to use for virtual memory, and changes to the boot flow to support the ZFS filesystem (discussed in~\Cref{sec:zfs}).

In addition to our experiment, Edge Node Lite supported a flight demonstration of the lake and island identification algorithm described in~\cite{carragher2024} to improve the efficiency of data collection on the spacecraft.

\section{METHODS}

\subsection{Problem Statement}

\begin{figure*}
    \centering
    \includegraphics[width=\linewidth]{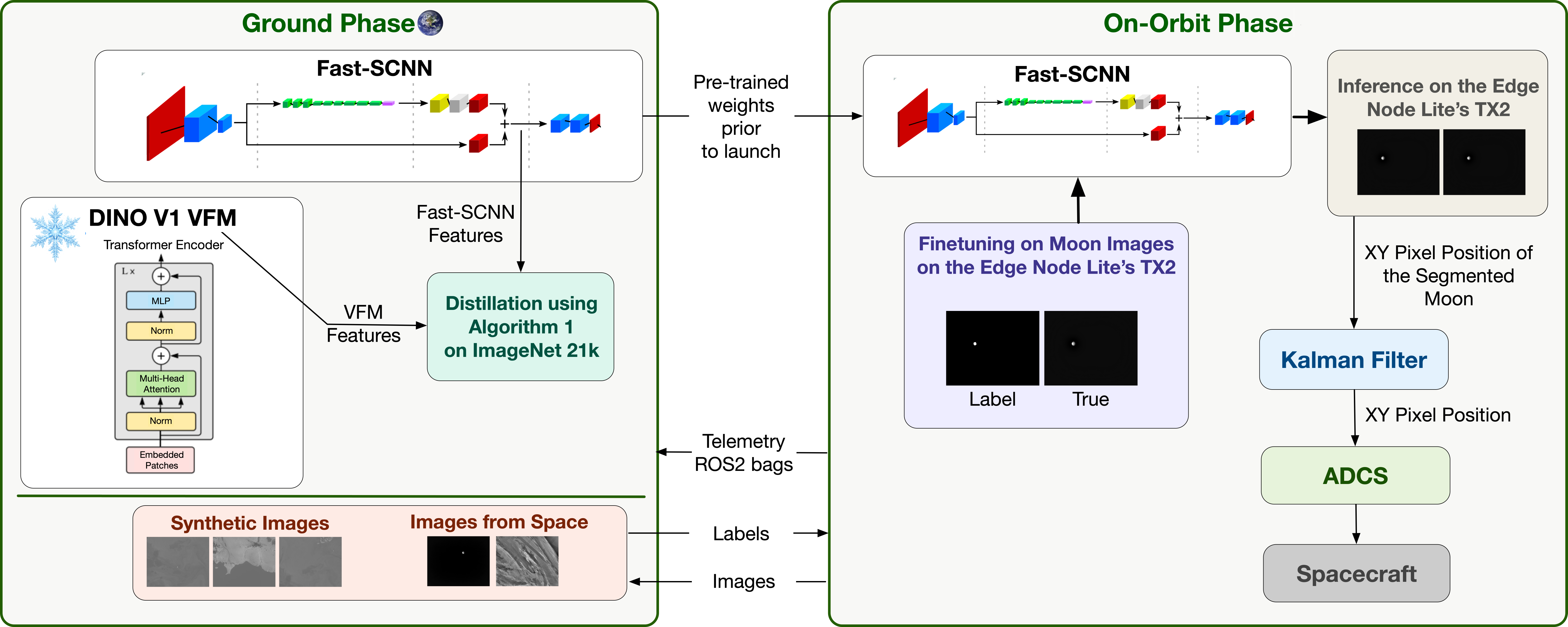}
    \caption{\textbf{Overview of our method.} On Earth, we perform distillation of the VFM into the Fast-SCNN network using~\Cref{alg:training_vision_satellite}. As interfaces, we downlink ROS 2 bags and other logs from the satellite, as well as images, and uplinked labels after annotation. 
    In space, we perform both training and inference of images captured by the onboard camera. We also close the loop on the segmented images to enable the spacecraft to point towards the Moon. The ``snowflake'' symbol represents the fact the DINO model is frozen (no gradients flow through the network).}
    \label{fig:methods}
\end{figure*}

We consider the problem of vision-based spacecraft attitude control in LEO.
Our approach is to use a learning-based perception model that is trained onboard the spacecraft, as seen in~\Cref{fig:methods}.
The solution consists of two components:
(1) the perception model (a segmentation network) is trained directly on the spacecraft using labels uploaded from Earth and onboard compute resources (the NVIDIA Jetson TX2 NX's GPU) and
(2) the spacecraft employs the onboard-trained segmentation model to estimate the Moon's position in the image stream and generates attitude commands to align (point) the spacecraft toward the Moon. We refer to this method as dynamic targeting throughout this paper.
Note that this work is intended as a technology demonstration of key components for onboard learning and control; alternative approaches to vision-based spacecraft control are possible and remain an important direction for future work.

\subsection{Approach}\label{sec:approach}

\subsubsection{Segmentation Module}\label{sec:segmentation_module}
Our proposed method for segmentation utilizes Fast-SCNN~\cite{poudel2019fast}, a CNN optimized for fast inference times on edge-compute devices.
The main components of the network are a learning-to-downsample module, a coarse global feature extractor, a feature fusion module, and a standard classifier.
The global feature extractor module captures the global context for image segmentation and the feature fusion module merges these features using addition, as shown in the Fast-SCNN architecture in~\Cref{fig:methods}.
Because real-world spacecraft datasets are limited in size, we pretrain Fast-SCNN in order to prevent overfitting.
Specifically, we do this by distilling a Visual Foundation Model (VFM), DINO V1~\cite{caron2021emerging}, via  student-teacher training using the ImageNet-21k dataset~\cite{5206848}. This dataset contains approximately 14.2 million images of different objects, including space-related ones.
After distillation is complete, we finetune the Fast-SCNN segmentation network on the dataset of images in orbit.

The distillation process (\Cref{alg:training_vision_satellite}) is as follows. Given an input image $\mathcal{I}\in \mathbb{R}^{H\times W \times 3}$, where $H$ is the image height, $W$ is the image width and $3$ represents the number of channels, we first perform a forward pass using DINO\footnote{DINO~\cite{caron2021emerging} is based on a vision transformer architecture, which operates on a sequence of $16\times16$ or $8 \times 8$ patches from the input image.}, our teacher network, and extract an intermediate feature map $\vb*{z}_{\textrm{vfm}} \in \mathbb{R}^{384 \times \floor{\frac{HW}{16\cdot 16}}}$ before reshaping it such that
\begin{equation}\label{eq:zvfm}
    \vb*{z}_{\textrm{vfm}} \in \mathbb{R}^{\floor*{\frac{H}{16}} \times \floor*{\frac{W}{16}} \times 384},
\end{equation}
where $\floor*{\cdot}$ denotes the floor operator, and 16 and 384 denote the patch size and feature size, respectively. We then extract the intermediate feature map from the student network, Fast-SCNN\footnote{For Fast-SCNN, we update the relevant convolution kernel to have an output channel dimension of 384 in order to produce feature maps that match the channel dimension of DINO's corresponding feature map.}, by performing another forward pass to get 
\begin{equation}
    \vb*{z}_{\textrm{cnn}} \in \mathbb{R}^{\floor*{\frac H 8} \times \floor*{\frac W 8} \times 384}.
\end{equation} 

To train the Fast-SCNN, we regress its feature map to that of DINO using the mean squared error loss function, updating only the weights of the Fast-SCNN via stochastic gradient descent (Lines~\ref{alg:loss}-\ref{alg:sgd}). Note that we only backpropagate with respect to Fast-SCNN's encoder weights, since the segmentation head is not involved in the computation of the loss. During the loss computation, we also upsample via bicubic interpolation the feature maps such that they have the same dimensions.

Lastly, we train the DINO-distilled Fast-SCNN on the segmentation task. Note that the segmentation task running onboard the spacecraft uses thermal images (1 channel), rather than RGB images (3 channels). 
To accommodate thermal images, the first convolutional layer, with shape $[\mathrm{out} \ \mathrm{channels}, 3, \mathrm{kernel} \ \mathrm{height}, \mathrm{kernel}  \ \mathrm{width}]$ is changed to $[\mathrm{out} \ \mathrm{channels}, 1, \mathrm{kernel} \ \mathrm{height}, \mathrm{kernel} \ \mathrm{width}]$ by averaging across the 3 RGB channels.

On the spacecraft, we demonstrate inference of our DINO-distilled Fast-SCNN as well as training (finetuning) onboard. 
The training parameters (momentum, optimizer, step size) are kept consistent with the initial Fast-SCNN paper~\cite{poudel2019fast}.

\algblock{Input}{EndInput}
\algnotext{EndInput}
\algblock{Output}{EndOutput}
\algnotext{EndOutput}
\begin{algorithm}[t]
    \caption{VFM distillation into a real-time CNN (run on the ground)}
    \label{alg:training_vision_satellite}
    \begin{algorithmic}[1]
        \State \textbf{Input:}  Networks $\mathcal{F}_\textrm{vfm}$, $\mathcal{F}_\textrm{cnn}$, training dataset $\mathcal{D}$
            \\ \qquad\quad Number of iterations N, step size $\eta$, \\ \qquad\quad VFM weights $\boldsymbol{\theta}_\mathrm{vfm}$, initial CNN weights $\boldsymbol{\theta}_\mathrm{cnn}$.
        \State \textbf{Output:} DINO-distilled CNN weights $\vb*{\theta}_\textrm{cnn}$. 
        \\
        \For{$n=1:N$}
            \State Sample image batch $\vb*{x_n}$ from $\mathcal{D}$.
            \State $\vb*{z}_{\textrm{vfm}} := \mathcal{F}_\textrm{vfm}(\vb*{x}_n ; \ \vb*{\theta}_\textrm{vfm}).$ \Comment{Get VFM features} \label{alg:z_vfm}
            \State $\vb*{z}_{\textrm{cnn}} := \mathcal{F}_\textrm{cnn}(\vb*{x}_n ; \ \vb*{\theta}_\textrm{cnn}).$ \Comment{Get CNN features} \label{alg:z_cnn}
            \State $\mathcal{L} := \Vert \vb*{z}_{\textrm{cnn}} - f_\textrm{bicubic}(\vb*{z}_{\textrm{vfm}}) \Vert_2^2.$ \label{alg:loss}
            \State $\vb*{\theta}_\textrm{cnn} := \vb* {\theta}_\textrm{cnn} - \eta  \nabla_{\vb*{\theta}_\textrm{cnn}} \mathcal{L}.$\label{alg:sgd} \Comment{Using Adam}
        \EndFor
    \end{algorithmic}
\end{algorithm}

\textbf{Baseline.}\label{sec:baselines}
As a baseline we used the Fast-SCNN without knowledge distillation. We used the baseline only for ground experiments on some of the datasets presented next, not for inference or training in space.

\begin{figure*}[t]
    \centering
    \includegraphics[width=\linewidth]{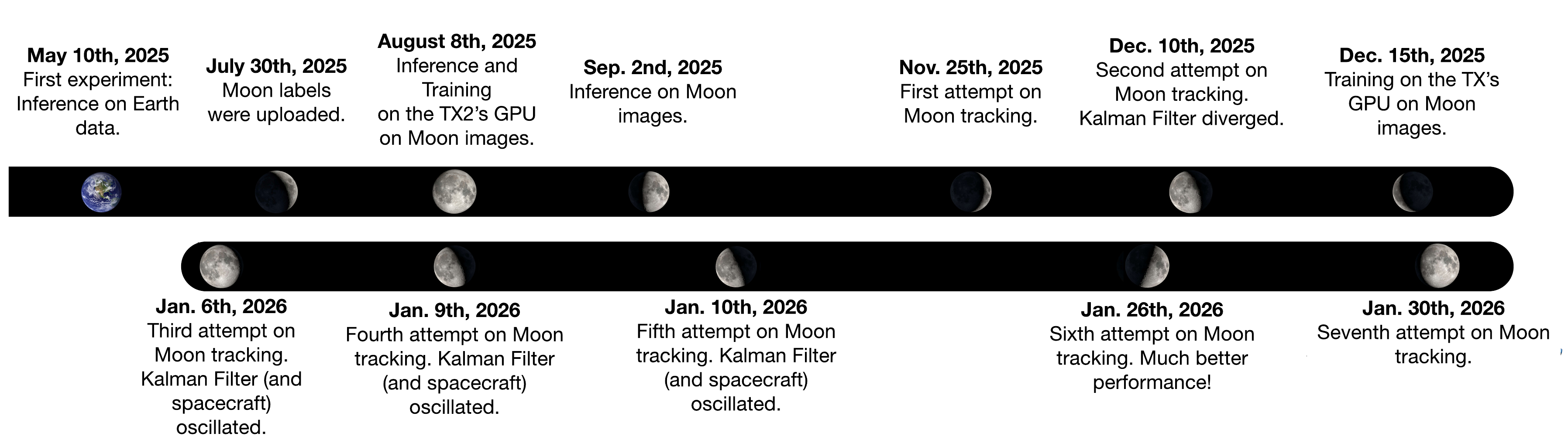}
    \caption{\textbf{Timeline of the different experiments run onboard the spacecraft since its launch.} In total, we ran 12 experiments, the majority being inference and dynamic targeting experiments. We have run three training experiments on the TX2 NX's GPU on Moon images with labels uploaded from the ground (Aug. 8th, 2025 and Dec. 15th, 2025). The weights from Aug. 2025 were used for the Sep., Nov., and Dec. experiments, and the ones from Dec. for the Jan. experiments. The Moon icon signifies in which phase the Moon was when the experiment was performed. See~\Cref{tab:params_experiments} for the parameters used in the Jan. 2026 tracking experiments. For discussion on these parameters, see \Cref{sec:dynamic_retargeting}.}
     \label{fig:experiments_timeline}
\end{figure*}


\textbf{Datasets.}
We validated DINO + Fast-SCNN (our method) and the baseline on two datasets of RGB images and one dataset of synthetically generated thermal data before deploying it to space. While in space, we collected additional data of the Moon for both inference and training.
More details on the design of these datasets and benchmarking are presented in~\Cref{sec:ground_experiments}.

\vspace{-0.6cm}

\subsubsection{Dynamic Targeting}\label{sec:closing_loop}

We close the loop between the ADCS  on the spacecraft and the segmentation module. 
As shown in~\Cref{fig:methods}, we first extract centroid as X and Y coordinates in the image plane. We then feed those coordinates to a Kalman Filter (KF), which outputs filtered X, Y coordinates. Those coordinates are then sent to the ADCS system, which orients the spacecraft to point towards the desired direction, in this case, the Moon.

We design the KF to estimate the position of the Moon in the image plane. 
Let $\mathbf{x} := [u, v, \dot{u}, \dot{v}]^\top \in \mathbb{R}^4$ be the state of the Moon in the image frame. 
We assume a constant velocity model for the translational component of the target motion in the image plane, which is reasonable for low slew rates. The dynamics are then
$
\dot{\mathbf{x}} = \mathbf{A}\mathbf{x} + \mathbf{w},
$
where $\mathbf{w}\in \mathbb{R}^{4}$ is the process noise and $\mathbf{A} \in \mathbb{R}^{4x4}$ is the dynamics matrix, as follows
$$
\mathbf{A} = \begin{bmatrix}
    0 & 0 & 1 & 0\\
    0 & 0 & 0 & 1\\
    0 & 0 & 0 & 0\\
    0 & 0 & 0 & 0
\end{bmatrix}.
$$
The measurement equation is
$
\mathbf{z}= \mathbf{H} \mathbf{x} + \mathbf{v},
$
where $\mathbf{z}\in \mathbb{R}^2$ is the measurement vector, $\mathbf{v}\in \mathbb{R}^2$ is the measurement noise, and $\mathbf{H}\in \mathbb{R}^{2 \times 4}$ is the measurement matrix
$$
\mathbf{H} = \begin{bmatrix}
1 & 0 & 0 & 0 \\
0 & 1 & 0 & 0
\end{bmatrix}.
$$
The KF resets if the covariance magnitude, computed as the square root of the trace, is larger than a threshold.

\section{RESULTS}\label{sec:results}

In this section, we introduce the results from 12 experiments run on the spacecraft (\Cref{fig:experiments_timeline}), as well as other results from our ground testing used to validate our solution.


For the majority of experiments, we report the mean Intersection over Union (mIoU) performance metric. The IoU for an individual class is defined as the area of overlap between the predicted segmentation and the ground truth mask, normalized by the total area covered by their union; the mIoU is calculated by averaging these values across all target classes. As such, an mIoU score of 1.0 indicates a perfect prediction, while a score of 0.0 indicates a complete mismatch.
During both training and inference, we also report the CPU/GPU usage and the thermal performance.
\vspace{-0.2cm}
\subsection{Training Results on Spacecraft}

\begin{figure}[t]
    \centering
    \includegraphics[width=\linewidth]{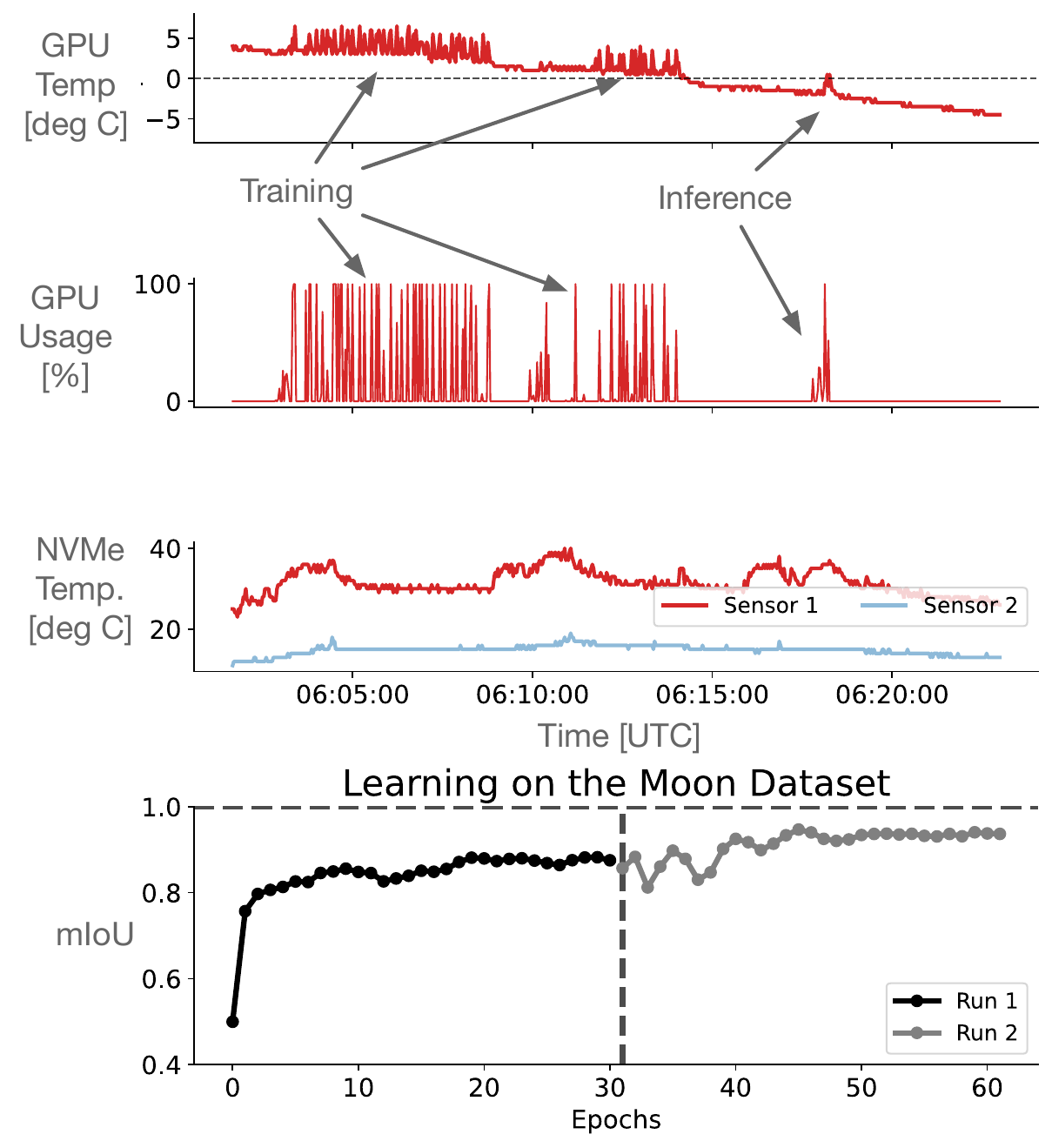}
    \caption{\textbf{Performance of the training and inference experiments in space (August 8th).} (top plot) Thermal results for the GPU. The experiment was run while the spacecraft was in eclipse. 
    (second plot) GPU usage during training and inference.
    (third plot) Thermal results for the NVMe. 
    For the first 3 plots, the data are sampled at 1~Hz.
    (bottom plot) Training running in two cycles for 60 epochs total. 
    }
    \label{fig:training}
\end{figure}

On August 8th and December 15th, we conducted three on-orbit training experiments of the DINO-distilled Fast-SCNN model on the NVIDIA Jetson TX2 NX. Training of the Fast-SCNN model was performed with a batch size of 2 and a 75\%/5\%/20\% train/validation/test split (see~\Cref{sec:ground_experiments} for details on the dataset splits). All the other parameters were kept the same as in~\cite{poudel2019fast}. 

The training was done using the labels that were generated on the ground and uploaded to the spacecraft. 
Although we collected many images of the Moon prior to the training experiment, we down-selected these to a representative set of 40 images (for example, images of the type seen in~\Cref{fig:architecture}) including different phases of the Moon and overexposed images.
Training was initialized from the loaded weights included before launch, which were obtained using~\Cref{alg:training_vision_satellite}. 
In~\Cref{fig:training}, we present the results of this training procedure. The bottom plot shows rapid performance improvement after the first epoch, followed by stabilization, reaching a final mIoU exceeding 0.9. The experiment consisted of 30 epochs, followed by an additional 30 epochs run, for a total training time of approximately 13 minutes.
We split training in 2 phases in order to also demonstrate finetuning from weights generated on-orbit, not just uploaded from the ground.
We stored the final model weights onboard and subsequently used them for inference experiments and dynamic targeting, as discussed in the next section.

Apart from the mIoU performance, in~\Cref{fig:training}, we show thermal performance of the GPU and NVMe during training and inference. Note that the spacecraft was in eclipse while the experiment was running. We observe GPU temperature spikes of about 1-4 degrees C during training. The overall temperature trend is generally decreasing as the spacecraft spends more time in eclipse.
The third plot in~\Cref{fig:training} shows the GPU usage, which is more pronounced during training rather than during inference, as expected.
There is a gap between the two training runs due to the time taken to stop and restart the Docker container (see~\Cref{sec:operations_dockers} for a description on the operations).

\begin{table}[t]
\centering
\caption{Parameters used for the closed-loop control experiments in the experiments performed in January 2026.}
\label{tab:params_experiments}
\resizebox{\columnwidth}{!}{
\begin{tabular}{p{2.5cm}ccccc} 
\toprule[1pt]
\makecell[l]{Experiment \\ Date} & \makecell{Boson \\ Frame Rate} & \makecell{KF Prediction \\ Rate} & \makecell{Control \\ Gain} & \makecell{Low-Pass \\ Filter} & \makecell{Boson Frame \\ Buffer} \\ 
\midrule[1pt]
Run 3: Jan. 6th, 2026 & 1 Hz & 10 Hz & 2 & None & 4    \\
Run 4: Jan. 9th, 2026 & 1 Hz & 10 Hz & 2 & None & 1  \\
Run 5: Jan. 10th, 2026 & 2 Hz & 2 Hz  & 1 & Enabled & 1   \\
Run 6: Jan. 26th, 2026 & 2 Hz & 2 Hz  & 1/2 & Disabled & 1    \\
Run 7: Jan. 30th, 2026 & 2 Hz & 2 Hz  & 1/2 & Disabled & 1   \\
\bottomrule[1pt]
\end{tabular}
}
\end{table}
\vspace{-0.2cm}
\subsection{Dynamic Targeting}\label{sec:dynamic_retargeting}

\begin{figure*}[t]
    \centering
    \includegraphics[width=\linewidth]{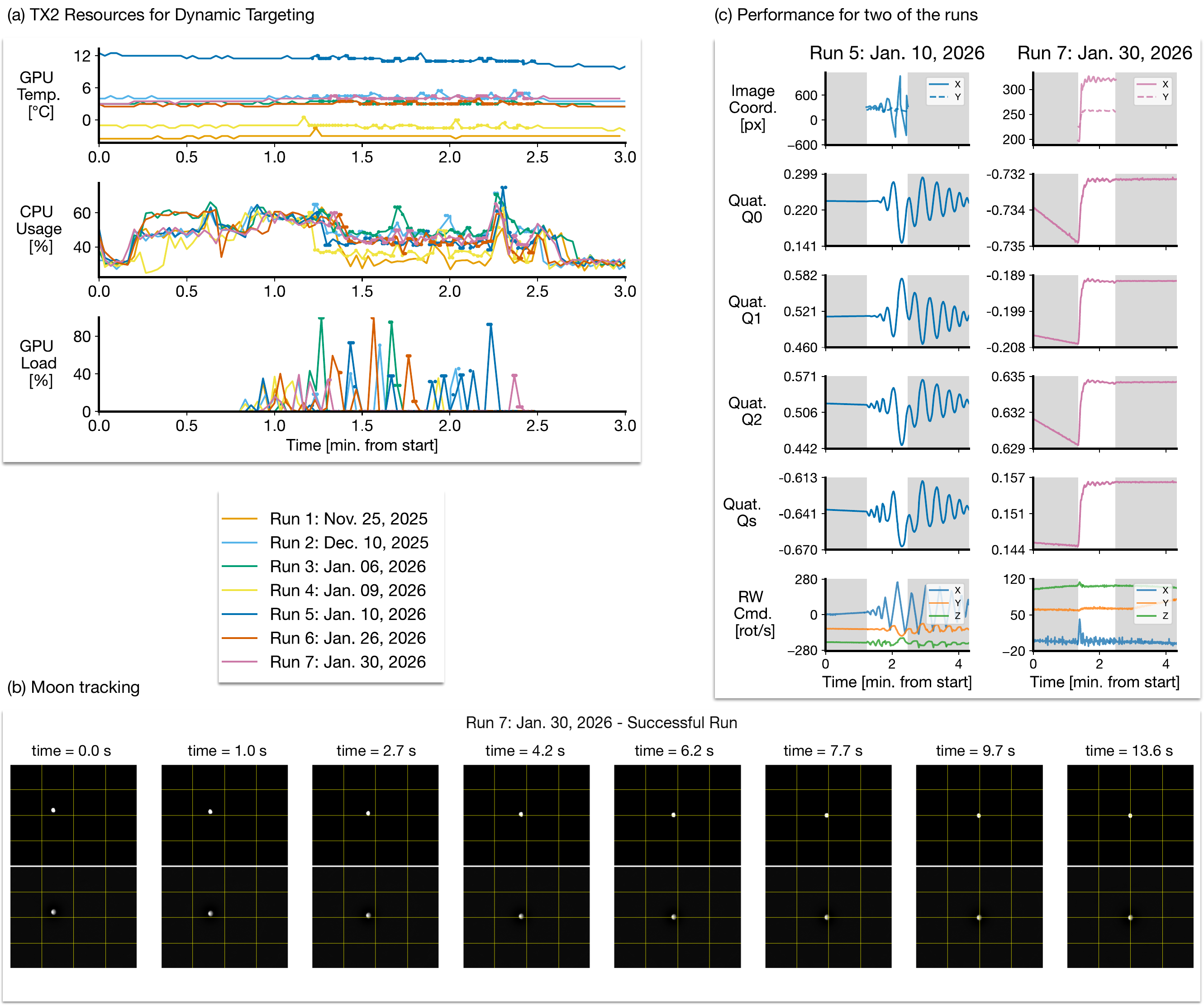}
    \caption{\textbf{Performance on the dynamic targeting task.} (a) Results from the dynamic targeting experiment runs. (top) GPU temperature, (middle) CPU usage in total over the 4 cores, (lower) GPU load. The dots show when image inference was performed. For the runs on and after Jan. 10th, we increased the inference frequency from 1 Hz to 2 Hz.
    (b) Position of the Moon during the dynamic targeting phase (first row contains a sequence of segmented Moon images and the second row represents a sequence of true thermal images). The Moon was moving towards the center of the image, showing that Run 7 performs well. 
    (c) Performance from two runs (image coordinates, quaternions, and reaction wheel commands). For a discussion on the cause of the oscillating behavior on Run 5, see \Cref{sec:dynamic_retargeting}.}
    \label{fig:pointing_runs}
\end{figure*}

We evaluated the closed-loop performance of the system in a set of dynamic targeting experiments, where the onboard-trained segmentation model was used in real time to send commands to the spacecraft (\Cref{fig:pointing_runs}). 
In these experiments, the Moon was continuously detected in the image stream using our segmentation module. 
The planar position of the Moon in the image plane was passed through a KF, introduced in~\Cref{sec:closing_loop}, and the output of the filter was used to generate attitude commands to reorient the spacecraft toward the target, as seen in~\Cref{fig:pointing_runs}b.
\Cref{fig:pointing_runs}a shows the compute characteristics during these experiments. The dots correspond to inference events, highlighting the bursty nature of GPU utilization (note that the recording rate was 1 Hz and the bursts were around 3-4 seconds).

Early experiments exhibited oscillation of the spacecraft in the horizontal plane (overshooting the Moon target rather than tracking it well), as seen in the quaternion commands and pixel coordinates in \Cref{fig:pointing_runs}c. However, this was greatly reduced in the two tracking experiments run on Jan. 26th, 2026 and Jan. 30th, 2026. We made the following changes to achieve stable tracking, which are summarized in~\Cref{tab:params_experiments}:
\begin{itemize}[noitemsep]
    \item We changed the frame buffer size used by OpenCV for the Boson camera from the default of 4 frames to 1 frame, which means that the frames being processed are closer to real time. This patch was applied for the Jan. 9th, 2026 run.
    \item We changed the frame rate from 1 Hz to 2 Hz for the Jan. 10th and later runs so that the feedback loop is tighter. 
    The KF prediction rate was also set to 2 Hz to match the frame rate. A higher frame rate and inference rate could be explored in future work; the camera supports 60 Hz.
    \item The control gain was reduced twice (on Jan. 10th and again on Jan. 26th). Before Jan. 10th, a pixel coordinate 10 pixels left of center would cause an attitude quaternion to be calculated to slew to a point 20 pixels left of center, due to the control gain being inadvertently set to 2. On the Jan. 26th run, a control gain of 0.5 was used so only a 5 pixel delta quaternion was commanded in case the KF output was 10 pixels off-center. This helped to reduce oscillation due to the latency in the feedback loop.
    \item For the Jan. 10th, 2026 run, we enabled a low-pass filter which smoothed the calculated rotation vectors (with a 4 second time window) before converting them to quaternion commands. However, this was disabled for the Jan. 26th and subsequent runs, as it was found to be counterproductive for fast target tracking with relatively low measurement noise. The filter would be more beneficial in scenarios with noisier measurements or slower required slew rates.
    \item The small remaining oscillations were caused by noise in the segmentation.
\end{itemize}

\begin{table}[t]
\centering
\caption{Quantitative results of the pointing accuracy for dynamic targeting experiments. Values are reported in pixels.}
\resizebox{\columnwidth}{!}{
\begin{tabular}{llllll}
\toprule[1pt]
Experiment Date        & Mean  & Final & $\Delta$err & $\Delta$X & $\Delta$Y \\ 
\midrule[1pt]
Run 2: Dec. 10, 2025 & 167.8         & 197.2          & 127.0                & 117.6              & 71.2               \\ 
Run 3: Jan. 06, 2026 & 199.7         & 188.1          & 612.9                & 970.0              & 278.9              \\ 
Run 4: Jan. 09, 2026 & 579.8         & 2356.6         & 2335.5               & 4444.1             & 71.0               \\ 
Run 5: Jan. 10, 2026 & 176.8         & 37.0           & 726.8                & 1481.7             & 82.4               \\ 
Run 6: Jan. 26, 2026 & 7.4           & 4.0            & 20.1                 & 34.1               & 6.8                \\ 
Run 7: Jan. 30, 2026 & 13.5          & 1.2            & 3.5                  & 6.9                & 3.7                \\ 
\bottomrule[1pt]
\end{tabular}
}
\label{tab:qualitative_results}
\end{table}

In~\Cref{tab:qualitative_results}, we report pointing accuracy for the six ADCS experiments over the first $3 \min$ of each run. 
The error $e$ is the radial distance in pixels from the commanded image point to the frame center at $(320 \mathrm{px},256 \mathrm{px})$. 
We report the `mean' as the average of $e$ over the first 3 minutes, the `final' as the last error in that window, and $\Delta \mathrm{err} = \mathrm{max}(e) - \mathrm{min}(e)$, $\Delta x=\max (x-320)-\min (x-320)$, $\Delta y=\max (y-256)-\min (y-256)$  
the errors
over the last 30 s of the first 3 minutes.
In some earlier experiments, we have not clamped the output of the KF within the boundaries of the image.
Lastly, Run 1 was excluded from~\Cref{tab:qualitative_results} because the code crashed.

Overall, these results demonstrate that dynamic, vision-based closed-loop control can be achieved onboard a spacecraft using modern learning-based perception models under realistic compute constraints.

\vspace{-0.2cm}

\subsection{Performance of the NVIDIA Jetson TX2 NX in LEO}\label{sec:performance_tx2_leo}
Six months after launch, and again 12 months after launch, we ran a ZFS scrub on the spacecraft to check every file for corruption. During normal operation, when a scrub is not underway, only files that are read are checked by ZFS for corruption, so corruption can go unnoticed because many files are unused. In our case, ZFS found no checksum errors, which can likely be attributed to (a) the radiation tolerance of the SSD design, detailed in~\Cref{sec:radiation_testing}, (b) the relatively benign radiation environment in LEO, and (c) the fact that the Edge Node Lite payload is powered off most of the time, and only on during experiments.

We record telemetry from the NVMe SSD including temperature, health status, and time spent powered on. After a year of use on-orbit, the SSD recorded approximately 45 hours of time powered on.

While we did not run any tests on-orbit that intentionally verify the integrity of GPU calculations on-orbit, we did not see any behavior in the course of running experiments that would indicate Single Event Effects or any other type of radiation induced effects.
\vspace{-0.2cm}

\subsection{Radiation Testing}\label{sec:radiation_testing}

\subsubsection{Solid State Drive (SSD)} In August 2022, The Aerospace Corporation conducted proton beam radiation testing of four NVMe SSDs to determine their fitness for use in LEO: Western Digital IX SN530 1 TB, Western Digital SN530 512 GB, Swissbit 320 GB (SFPC320GM1AG4TO-I-8C-51P-STD), and Greenliant 960GB (GLS88BP960G3). The testing was conducted at Loma Linda University Medical Center's High-Energy Beam Line from August 5-6, 2022, with 202 MeV protons at the sample location over approximately 8 hours of cumulative beam exposure. Seven total devices were tested across the four models from three vendors (Greenliant, Swissbit, and Western Digital). Each SSD was installed in a USB3 NVMe enclosure connected to a dedicated NVIDIA Jetson Nano single-board computer, which performed continuous write-read-verify cycles using alternating bit patterns (0xAA and 0x55) while exposed to proton radiation. The test framework monitored throughput performance and data integrity in real-time, with automatic failure detection for stalled I/O or significant throughput degradation. Results showed the drives transacted substantial data volumes during exposure: the WD IX SN530 1TB drives completed 3,323 GB and 575 GB respectively, the WD SN530 512GB completed 2,004 GB, the Swissbit 320GB devices completed 968 GB and 65GB, and the Greenliant 960GB devices completed 1,252 GB and 124 GB.

The testing revealed that all WD devices suffered 10 times fewer Single Event Function Interrupt (SEFI) faults than the Greenliant or Swissbit drives, with one destructive event permanently destroying a Greenliant device. SEFI faults require a power cycling of the host NVIDIA Jetson computer to restore NVMe operations. For an orbit in a benign radiation environment such as approximately 500 km, the expectation for the SN530 is approximately 0.5 SEFI per year. Destructive events, estimated at~0.05 events per year (10 times lower than SEFI), translate to approximately 10\% probability of permanent device failure after 2 years on orbit. Based on these results, the recommendation is to fly the WD IX SN530 1TB NVMe SSD for LEO applications. However, several important limitations apply: data corruption during testing was rare but hints were observed (a few byte corruptions detected when using smaller 1GB test blocks rather than the original 100GB blocks); the test did not comprehensively assess total ionizing dose (TID) effects and additional Cobalt-60 testing is recommended to more robustly understand TID behavior; USB power consumption could not be measured to determine if devices are latchup-free; and concepts of operations must be adapted to accommodate the SEFI characteristics requiring periodic power cycling.

These tests accessed the NVMe drives directly as block devices, so that the filesystem was not a factor in the results and only the drive operation was tested. Testing was conducted before we had implemented support for OpenZFS, described in~\Cref{sec:zfs}.

\vspace{-0.5cm}

\subsection{Implementation Details}
\subsubsection{ROS 2 Middleware}
Our Docker images included ROS 2 Foxy with the Fast DDS implementation of the DDS protocol, configured to use shared memory for communication between processes. We included ROS 2 topics for image frames, Jetson resource utilization statistics, spacecraft telemetry such as temperature readings, attitude, and position information. We provided ROS services for attitude commanding by geographic coordinates or raw quaternions. Pixel-based attitude commanding through ROS 2 is also supported, and is the method used by the work described in this paper.

The supervisor Microcontroller on Edge Node Lite (SAMV71Q21RT) runs a Real-Time Operating System and includes the micro-ROS codebase, which allows it to participate in the ROS network running on the Jetson TX2 NX. The Jetson TX2 runs a bridge software called micro-ROS Agent, which bridges between standard DDS (used by ROS 2) and Micro Extremely Resource Constrained Environments DDS (Micro-XRCE-DDS), a lightweight DDS variant for microcontrollers, communicating with the Microcontroller using UDP packets. The Microcontroller is the interface between the avionics stack and the Jetson TX2, and thus is able to publish state of health telemetry from the avionics stack to a ROS topic, make Jetson state of health telemetry available directly to the Flight Computer, and provide a ROS service to pass along attitude commanding from the Jetson to the Attitude Control Board.

\vspace{-1.0cm}

\subsubsection{File System (ZFS)}\label{sec:zfs}
We used OpenZFS, which is an open-source implementation of the ZFS file system and volume manager, that supports the Linux kernel. Significant work was required to integrate ZFS with NVIDIA's Linux for Tegra software stack (which uses Extended 4 by default) and the bootloader U-Boot, including patches that were upstreamed to U-Boot as a result of this work.

ZFS is a filesystem that has decades of use in production systems terrestrially, but, to the best of our knowledge,  had not been used in space. The primary characteristics that make ZFS appealing for space systems are its robustness to sudden power loss (due to its copy-on-write architecture) and its layers of redundancy and checksums to detect and repair data corruption, both for filesystem metadata and file content. ZFS is able to detect and repair corruption of metadata or file content using in-memory copies, on-disk copies, or copies distributed across multiple drives (although this latter RAID-Z capability was not used on this mission).

In our case, we configured ZFS to duplicate critical Operating System files using the ``copies=2'' setting, meaning two copies are kept for each of those files, whereas sensor data files were not duplicated (to conserve disk space). We also utilized the snapshot feature of ZFS to store a snapshot of the filesystem at the time of launch, with a mechanism to trigger a reset to that state via a GPIO pin controlled by the microcontroller, in case there arises a need to revert to a known good state.

As space systems are vulnerable to both sudden power loss and data corruption due to radiation, ZFS is uniquely suited to this application.

\vspace{-0.7cm}
\subsubsection{Codebase} The codebase for our segmentation experiment is implemented in Python 3, while the experiment steps are orchestrated using Bash scripts.

\subsection{Ground Experiments}\label{sec:ground_experiments}
\subsubsection{Datasets}

\begin{figure}[t]
    \centering
    \includegraphics[width=\linewidth]{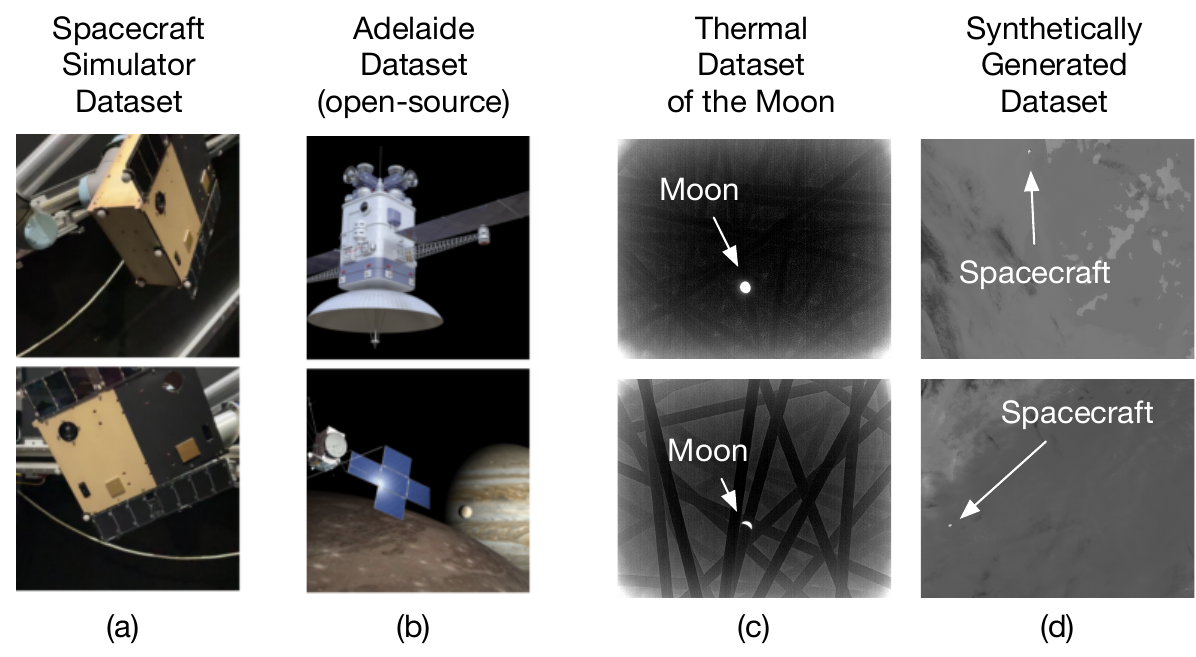}
    \caption{Example images from four datasets. The first dataset (a) consists of images taken inside our spacecraft simulator, while the second (b) is an open-source collection of spacecraft imagery~\cite{hoang_spacecraft_2021}. The remaining two datasets (c, d) are thermal: one includes in-orbit observations of the Moon, and the other contains synthetically generated images based on Landsat data with overlaid spacecraft.
}
    \label{fig:dataset_example_images}
\end{figure}

To validate our segmentation module (\Cref{sec:approach}) on the ground, we created several datasets.
One dataset consisted of segmenting spacecraft parts at short range. For this, we took RGB images of the Edge Node spacecraft in our spacecraft robotic simulator~\cite{nakka2018six} at different orientations and distances. Examples of such images are shown in~\Cref{fig:dataset_example_images}a.
We further utilized a publicly-available dataset consisting of 3117 images of various satellites and space stations~\cite{hoang_spacecraft_2021}.
This dataset included both synthetic and real images and videos, providing a diverse range of spacecraft types, as seen in~\Cref{fig:dataset_example_images}b.
For the thermal dataset, we used thermal images taken from the Landsat 8/9 satellites, which are equipped with sensors that fall within the LWIR range of the electromagnetic spectrum. These Landsat images provide the Earth background for the dataset, and vary in location, cloud coverage, and solar angle. We also used a FLIR Boson+ LWIR model 22640A012--6IAAX camera to capture images of the Edge Node spacecraft and remove the background (\Cref{fig:dataset_example_images}d). More details about this dataset can be found in~\cite{grauer2024visionbaseddetectionuncooperativetargets}. 
The last dataset was the Moon dataset, containing thermal images of the Moon captured on the spacecraft (\Cref{fig:dataset_example_images}c).


\vspace{-0.5cm}

\subsubsection{Performance on the RGB datasets on the ground}
In this section, we discuss the performance on the RGB datasets before flight.
We trained the DINO-distilled Fast-SCNN and the baseline (\Cref{sec:baselines}) on an NVIDIA Titan RTX GPU for a minimum of \texttt{100} epochs and selected the best performing epoch based on the mIoU of the validation dataset. 
We used the Adam~\cite{Kingma2014AdamAM} optimizer with a learning rate of \texttt{1e-3} and a batch size of \texttt{32}.
The results for training the models on the 75\%-20\%-5\% train, test, validation dataset splits are presented in Tables~\ref{tab:opensource_metrics_in} and \ref{tab:SR_metrics_in}.

On our custom spacecraft part segmentation dataset, we see that all models exhibit similar performance and nearly reach 1.0 (\Cref{tab:SR_metrics_in}). Interestingly, we found no benefit of model distillation in this dataset setting. This observation as well as the general high level of performance likely results from the smaller domain of our custom dataset. More details on this analysis are found in~\cite{grauer2024visionbaseddetectionuncooperativetargets}.

\begin{table}[t]
\centering
\caption{Spacecraft part segmentation performance (mIoU) on the Adelaide dataset.}
\label{tab:opensource_metrics_in}
\resizebox{\columnwidth}{!}{
\begin{tabular}{p{4cm}ccccc} 
\toprule[1pt]
Model & All & Body & Solar Panel & Antenna & Background \\ 
\midrule[1pt]
Fast-SCNN &0.683 & 0.705 & 0.707 & 0.347 & 0.974 \\
 DINO-distilled + Fast-SCNN (ours) & 0.756 & 0.764 &  0.768 & 0.480 & 0.982 \\
\bottomrule[1pt]
\end{tabular}
}
\end{table}

\begin{table}[t]
\centering
\caption{Spacecraft part segmentation model performance (mIoU) on our custom spacecraft simulator dataset.}
\label{tab:SR_metrics_in}
\resizebox{\columnwidth}{!}{
\begin{tabular}{p{4cm}cccc} 
\toprule[1pt]
Model & All & Body & Solar Panel & Background \\ 
\midrule[1pt]
Fast-SCNN &  0.971 & 0.968 & 0.958 &  0.986 \\
 DINO-distilled + Fast-SCNN (ours) & 0.972  & 0.969 & 0.961 &  0.987 \\
\bottomrule[1pt]
\end{tabular}
}
\end{table}

\subsubsection{Model Distillation Ablation Study}
In order to quantify the effect of our model distillation efforts for Fast-SCNN, we examine how segmentation performance is affected when fewer images are available for model training. Specifically, we examine the results when using only 75\%, 50\%, and 12.5\% of the total training set, with 5\% of the total data as validation set and 20\% as testing set.
\begin{table}[t]
\centering
\caption{Effect of training set size (\% of overall training set) on short-range spacecraft part segmentation (mIoU) for the custom dataset.
}
\label{tab:SR_split}
\resizebox{\columnwidth}{!}{
\begin{tabular}{p{4.5cm}  c c c} 
\toprule[1pt]
\multirow{2}{*}{Model} & \multicolumn{3}{c}{\% of Training Data Used} \\ \cmidrule(lr){2-4} 
 & 75\% & 50\% & 12.5\% \\ 
\midrule[1pt]
Fast-SCNN & 0.971 & 0.966&  0.933 \\
DINO-distilled + Fast-SCNN (ours) & 0.972 &  0.969 & 0.947 \\
\bottomrule[1pt]
\end{tabular}}
\end{table}
We first perform this ablation study on our custom spacecraft parts dataset, as well as on the Adelaide dataset. \Cref{tab:SR_split} and especially~\Cref{tab:opensource_split} show that the Fast-SCNN and DINO-distilled + Fast-SCNN model perform better with larger training datasets. 
In~\cite{grauer2024visionbaseddetectionuncooperativetargets}, we compare our method and the baseline against other common segmentation models such as YOLO.

\begin{table}
\centering
\caption{Effect of training set size (\% of overall training set) on short-range spacecraft part segmentation (mIoU) for the Adelaide dataset. }
\label{tab:opensource_split}
\resizebox{\columnwidth}{!}{
\begin{tabular}{p{4cm} c c c} 
\toprule[1pt]
\multirow{2}{*}{Model} & \multicolumn{3}{c}{\% of Training Data Used} \\ \cmidrule(lr){2-4} 
 & 75\% & 50\% & 12.5\% \\ 
\midrule[1pt]
Fast-SCNN &  0.683 & 0.598 & 0.537\\
DINO-distilled + Fast-SCNN (ours) & 0.756 & 0.722 & 0.636\\
\bottomrule[1pt]
\end{tabular}
}
\end{table}

\vspace{-0.5cm}
\subsubsection{Compute results}
We benchmarked the inference time of the short range detection models on the ground in Table~\ref{tab:hardware_metrics}.
We report results on the NVIDIA Jetson TX2 NX since it is the same computer as the one used on Edge Node Lite. We also provide benchmarks on an NVIDIA Titan RTX (4608 CUDA cores) for comparison to a workstation GPU.

\begin{table}[t]
\centering
\caption{Inference time (ground results) on two different platforms.}
\label{tab:hardware_metrics}
\resizebox{\linewidth}{!}{
\begin{tabular}{p{4cm} p{2.5cm} ccc}
\toprule[1pt]
\multirow{2}{*}{Model} & \multirow{2}{*}{Task} & \multicolumn{2}{c}{Inference Time (ms)} & \multirow{2}{*}{\makecell{Params \\ (M)}} \\ \cmidrule(lr){3-4} 
                   & & TX2 & Titan RTX &  \\ 
\hline
    \midrule[1.0pt]
Fast-SCNN & Semantic Segm. & 49.39  &  3.54 & 1.13 \\
DINO-distilled + Fast-SCNN (ours) & Semantic Segm. & 92.56 & 4.29 & 1.55\\
\bottomrule[1pt]
\end{tabular}
}

\end{table}

\section{DISCUSSIONS}

In this section, we discuss several findings and lessons learned from training vision-based ML models in space.

\subsection{Algorithm Performance}

Overall, the vision-based segmentation and dynamic targeting system demonstrated strong on-orbit performance.
Although Moon segmentation may appear straightforward, it presented several practical challenges in orbit, including image streaking, overexposure, and varying illumination conditions. These effects introduced noise and variability that complicated consistent segmentation performance.

More broadly, these experiments highlight a limitation of supervised learning approaches in space. While our use of knowledge distillation with Fast-SCNN allowed us to operate with a relatively small labeled dataset, reliance on labeled data is restrictive for more complex tasks, which is the ultimate goal of sending robots to explore space. 
Still, uploading labels is significantly less data heavy than uploading weights (i.e., 172 KB vs. 6.5 MB).
In many space applications, acquiring labels is impractical. 
Ideally, onboard systems should be capable of autonomous adaptation, operating without ground-provided supervision.

Finally, while the overall system performed well, there are clear paths for improvement.
The segmentation output exhibited imperfect boundaries between the Moon and the background, leading to variations in the estimated Moon centroid.
This, in turn, introduced measurement noise into the dynamic targeting pipeline, manifesting as small oscillations in the spacecraft pointing behavior. Improved segmentation accuracy, robust out-of-distribution detection, and further tuning of the Kalman filter are expected to reduce these effects and enhance overall system stability.

\subsection{Ground Testing Findings}
\subsubsection{Brown-out during inference and training}

When running our training algorithm on an engineering model, after about 30 seconds of operation, the NVIDIA Jetson TX2 experienced sudden power loss (brown-out).
We investigated the issue and the root cause was current spikes above the amount the flight power system can support, which resulted in a voltage droop below the 12V required for operation.

Earlier in development, for thermal and power testing, we used a matrix multiplication in an infinite loop to stress the GPU, assuming this would be representative of the flight algorithm which trains a neural network on the GPU. However, while the matrix multiplication has a higher average power consumption than the flight workload, the latter has very short spikes of current consumption above that average, which turned out to be more than the vehicle's power system can support.

To resolve the issue, we changed the Jetson's power mode from 20 Watts to 15 Watts. The power modes are provided by NVIDIA and affect the clock speeds of the CPU, GPU, and RAM.

\vspace{-0.5cm}
\subsubsection{ROS 2 Bag Compression}

The data from the experiments were recorded in a ROS 2 bag and downlinked to Earth for post-processing.

During our ground testing, we observed that the latency between issuing a signal interrupt and the completion of bag finalization is significantly impacted by enabling bag compression. 
For a representative 5 GB bag file, compression to approximately 4 GB required on the order of 10 minutes before the metadata file was generated and the recording process fully terminated. In contrast, for smaller bag files or when compression was disabled, finalization occurred almost immediately after signal interrupt.
This delay is non-negligible in an operational setting, particularly when timely access to recorded data is required. There is also a risk of the ROS 2 Bag's database being in an uncommitted Write-ahead Log state, which makes recovering the data therein difficult.
Based on these observations, we elected to disable compression during data recording and instead perform compression as part of the downlink preparation pipeline.

\subsection{Risk Tolerant Architecture Design}

The Edge Node Lite payload and the spacecraft bus avionics were designed as risk-tolerant platforms for experimental software and hardware payloads, with layers of safeguards built in to protect the vehicle in case of off-nominal operations. The system includes the following safeguards:
\vspace{-0.5cm}
\subsubsection{Payload-Level Safeguards}
\begin{itemize}[nosep]
    \item The Edge Node Lite payload's power is controlled by a Flight Computer command, and the power-on time is typically limited to under an hour. This means that, for example, the payload cannot inadvertently run for longer than intended.
    \item The Edge Node Lite payload's microcontroller controls power to the Jetson TX2, so even in a case such as the Jetson no longer booting successfully, we could still gather basic telemetry to diagnose the situation.
    \item Experiments on Edge Node Lite run in Docker containers with limited access to the filesystem.
    \item The service on the Jetson providing shell command execution capability to the bus runs on 2 dedicated CPU cores so even if the other CPU cores are over-subscribed command execution continues to function nominally.
\end{itemize}
\vspace{-0.6cm}
\subsubsection{Flight Computer Watchdog}
\begin{itemize}[noitemsep]
    \item The Flight Computer includes a hardware watchdog peripheral built into the VORAGO chip that automatically resets the device if the watchdog is not serviced as expected.
    \item The watchdog must be ``pet'' (serviced) by the lowest priority idle task at least once every 10 seconds, or it will trigger a reset. This ensures that if higher priority tasks stall and prevent the idle task from running, the watchdog will detect the problem.
\end{itemize}
\vspace{-0.6cm}
\subsubsection{Software Resilience}
\begin{itemize}[noitemsep]
    \item The Flight Computer maintains a read-only ``golden image'' containing the bootloader and base flight software that boots on every power reset. This golden image is not modified during flight.
    \item A secondary patchable image can be uploaded and used for software updates, but if it fails, the system automatically reverts to the golden image on the next boot.                    
\end{itemize}
\vspace{-0.6cm}
\subsubsection{Power Management and Safe Modes}
\begin{itemize}[noitemsep]
    \item The spacecraft's Flight Computer may enter safe mode depending on state of health conditions, with thresholds set for battery voltages, battery temperatures, and avionics temperature. This provides for controlled load shedding of non-vital subsystems to prevent no/low-power conditions.
    \item In safe mode, payloads are powered off, avionics run in idle mode, and the scheduler is disabled.
    \item Main power reset capability can power cycle or reset all avionics (except a few critical always-on components such as the Flight Computer and some core sections in the power subsystem) to recover from faults.
\end{itemize}
\vspace{-0.6cm}
\subsubsection{Communication Resilience}
\begin{itemize}[noitemsep]
    \item The spacecraft's antennas are nearly omni-directional, so in any attitude, communication with the ground is still possible.
    \item There are no attitude requirements for downlinking data, providing operational flexibility even during anomalies.
\end{itemize}



\subsection{Camera Streaks}\label{sec:camera_streaks}

As evident in~\Cref{fig:dataset_example_images}c, images taken with the FLIR Boson+ camera in space exhibit streak patterns, to varying degrees. 
These posed a challenge for effective image processing.
The streaks are believed to be caused by the Sun passing through the image frame as the spacecraft is slewing. We saw a modest reduction in streak patterns when we increased the camera's cadence of Flat Field Correction (FFC), which is a feature built into the firmware designed to correct for such patterns.

FFC works by periodically capturing a reference frame while the camera's shutter briefly closes, blocking all incoming light. This creates a ``dark frame'' that captures the camera's internal thermal noise and pixel non-uniformities. 
The camera then subtracts this reference pattern from subsequent images to remove fixed-pattern noise and correct for sensor variations caused by temperature changes and uneven detector response. In thermal cameras like the FLIR Boson+, this is particularly important because the sensor itself generates thermal noise that can create artifacts in the imagery.

\subsection{Operations}\label{sec:operations_dockers}

The system was designed to be highly streamlined and modular. Each component ran in its own Docker container, orchestrated via Docker Compose, which allowed us to manage the full multi-container system efficiently. Containerization means we can easily support different applications having different sets of dependencies, and allows limiting access to the host system. 
Deploying updates was simple: we only needed to upload a new source package and build it into an image onboard. 
In most cases we built artifacts from source during the Docker image build, although copying built artifacts would be more resource efficient and may be employed more widely on future missions. It would also be possible to build the Docker image layers on the ground and copy only new layers to the vehicle, but we are not aware of existing tooling for this, since most container tooling assumes an Internet connection is available or an image can be transferred in its entirety.

Using ROS 2 further improved the workflow. We could easily download recorded ROS 2 bags and rely on standard ROS 2 tooling for playback and analysis, enabling rapid debugging and iteration.
Importantly, our problem formulation minimized the need for heavy data transfers. Instead of uploading new model weights, we only needed to send lightweight labels, making the overall process significantly more efficient and easier to operate.

\section{CONCLUSIONS}

This work presents a flight demonstration of segmentation-based, closed-loop spacecraft autonomy on a 6U CubeSat, in which a deep learning perception model is trained onboard using LWIR images. We make the following contributions.

First, we demonstrate that distillation from a VFM yields good segmentation performance despite training on a limited set of Moon images captured in orbit.
Second, we conduct 12 experiments of onboard training and inference on a NVIDIA Jetson TX2 NX GPU in LEO, validating perception-based machine learning under space conditions, and demonstrating sufficient performance for closed-loop control.

Third, we develop a risk-tolerant autonomy software stack designed to accommodate experiments built on ROS 2 with Fast DDS and Micro-XRCE-DDS for data transport, and Docker-based containerization for ease of deployment.

We show thermal telemetry and CPU/GPU usage during training and inference, and discuss challenges encountered during ground testing.
The lessons learned from this mission show both the promise and the challenges of deploying learning-based autonomy in space.

\section*{ACKNOWLEDGMENT}
We thank S. Binkin for his contributions to the texturing of the Edge Node spacecraft in Blender, and S. Chien for his insightful suggestions on prior work in AI for space applications. On the Aerospace team, we thank G. Kinum for his work on ground integration testing, running operations on-orbit, and insights on issues such as the camera streaks. We appreciate E. Balerud's mechanical engineering work on the spacecraft, C. Day's electrical engineering design of the Edge Node Lite payload PCB, and N. Moreno's project management support.

We dedicate this mission to R. Lee of The Aerospace Corporation, whose vision and leadership were instrumental in its development prior to his passing in 2023.

\bibliographystyle{IEEEtran}
\bibliography{bibliography}

\input{bios}

\vfill\pagebreak

\end{document}

%% file: bios.tex
\begin{IEEEbiography}[{\includegraphics[width=1in,height=1.25in,clip,keepaspectratio]{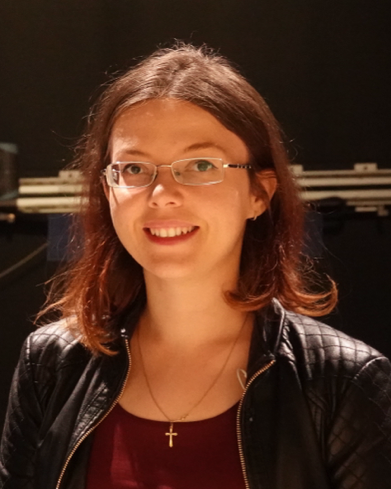}}]{Elena Sorina Lupu}~received her Ph.D. in Space Engineering from the California Institute of Technology (Caltech) where she is also affiliated with the Keck Institute for Space Studies. Currently, she is a Research Fellow at the Openmind Research Institute.
She has master's degrees from Caltech in Space Engineering and from the Swiss Federal Institute of Technology, Lausanne (EPFL) in Robotics and Autonomous Systems. Her current research focuses on autonomy, control, and reinforcement learning. 
She has won paper awards and fellowships including the AI4Science Fellowship and the Google Women Techmakers Scholarship, formerly Anita Borg.
\end{IEEEbiography}

\begin{IEEEbiography}[{\includegraphics[width=1in,height=1.25in,clip,keepaspectratio]{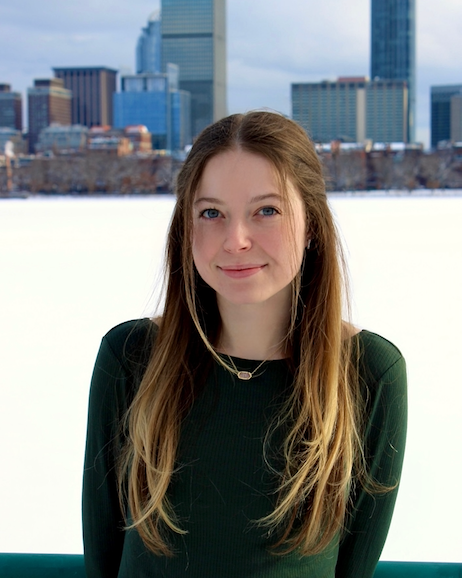}}]{Hannah Grauer}~is a PhD candidate in the MIT Center for Computational Science \& Engineering working in the Earth Intelligence Lab, where she is supported by the MIT x GE Vernova Climate Alliance Fellowship. She earned her B.S. in electrical engineering from Caltech. Her current research interests include computer vision and geospatial foundation models for remote sensing applications.
\end{IEEEbiography}

\begin{IEEEbiography}
[{\includegraphics[width=1in,height=1.25in,clip,keepaspectratio]{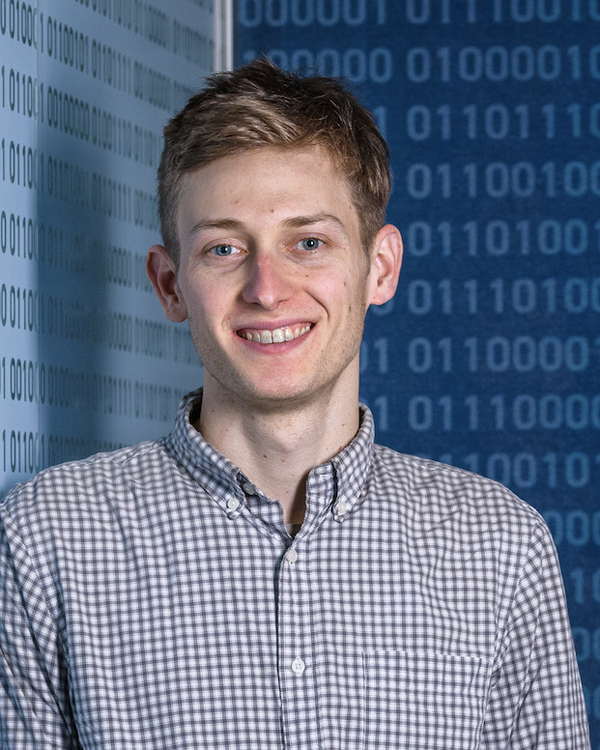}}]{Phaedrus Leeds}~earned his B.S. in Computer Science from The University of Alabama in 2017. As a software engineer at Endless, a nonprofit expanding access to educational technology in the Global South, he led the development of offline updates technology and contributed extensively to open-source software across the Linux Desktop stack. Currently at The Aerospace Corporation's Small Satellite Department, Phaedrus leads edge computing initiatives for space applications. Phaedrus collaborates with teams across government, industry, and academia to enable flight demonstrations of their technologies on CubeSat platforms.
\end{IEEEbiography}

\begin{IEEEbiography}
[{\includegraphics[width=1in,height=1.25in,clip,keepaspectratio]{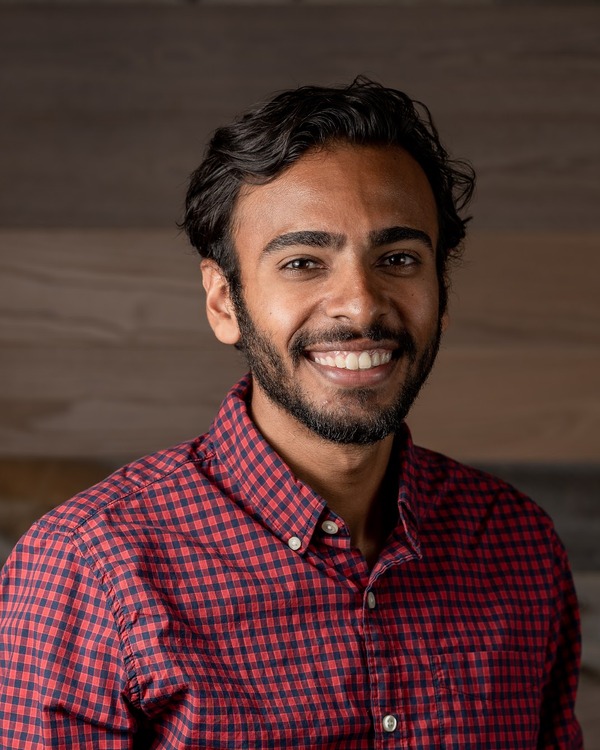}}]{Nikhil Ranganathan}~is a PhD candidate in the Aerospace Department at the California Institute of Technology. He earned his B.S. in Mechanical Engineering from Cornell University.  His current research interests include learning-based planning and control with applications to safety-critical robotic systems.
\end{IEEEbiography}

\begin{IEEEbiography}[{\includegraphics[width=1in,height=1.25in,clip,keepaspectratio]{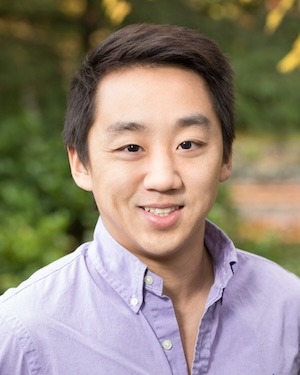}}]{Connor Lee}~ holds a Ph.D. in Space Engineering and a B.S. in Computer Science from the California Institute of Technology, received in 2024 and 2017, respectively. He was the recipient of the NSF Graduate Research Fellowship and a Best Paper Award Finalist, and has authored multiple publications in top robotics and computer vision venues. His research interests include learning-based perception for robotics, with applications towards aerial field robotics, autonomous driving, and vision-based navigation.
\end{IEEEbiography}

\begin{IEEEbiography}
[{\includegraphics[width=1in,height=1.25in,clip,keepaspectratio]{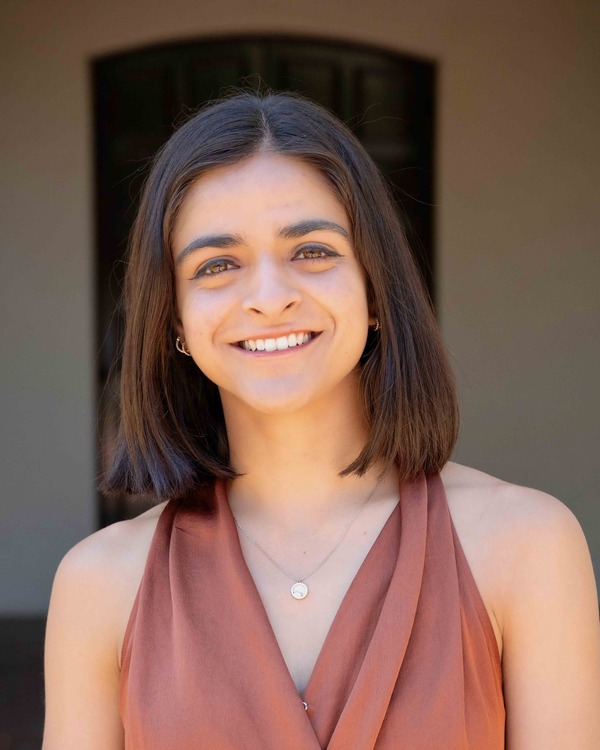}}]{Sulekha Kishore}~is a PhD student at the MIT Institute for Data, Systems, and Society (IDSS), supported by the NSF Graduate Research Fellowship Program. She earned her B.S. in Computer Science and Political Science from Caltech. Her current research focuses on large language models, AI for collective decision making in deliberative settings, and the societal impacts of AI. 
\end{IEEEbiography}

\begin{IEEEbiography}
[{\includegraphics[width=1in,height=1.25in,clip,keepaspectratio]{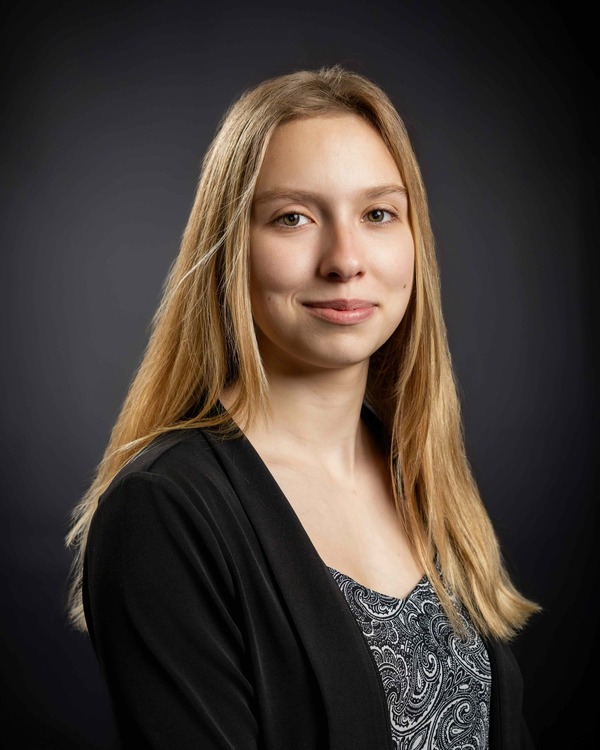}}]{Cailyn Smith}~is a PhD student in the Robotics Institute at Carnegie Mellon University. She holds a B.S. in Computer Science from Colorado School of Mines. She was a recipient of the 2025 NSF Graduate Research Fellowship. Previously, she interned at the Autonomous Robotics and Control Lab as a Caltech WAVE Fellow. Her current research interests include robot learning, human-robot interaction, and robotic manipulation.
\end{IEEEbiography}

\begin{IEEEbiography}[{\includegraphics[width=1in,height=1.25in,clip,keepaspectratio]{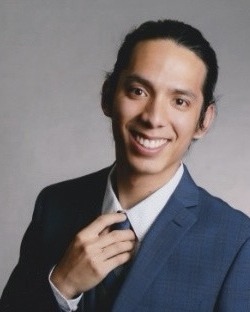}}]{Shon Cortes}~received his B.S. in Mechanical Engineering from California State University Fullerton and his M.S. in Robotics Engineering from University of Maryland College Park. Shon has 8 years of experience designing and building CubeSats and is a Member of Technical Staff in the Small Satellite Department at The Aerospace Corporation. His areas of focus are Software Engineering, Robotics, and Embedded Programming. He has primarily been developing the ROS Architecture, baseline sensor interface and ACS commanding nodes for The Aerospace Corporation’s CubeSat missions. 
\end{IEEEbiography}

\begin{IEEEbiography}
[{\includegraphics[width=1in,height=1.25in,clip,keepaspectratio]{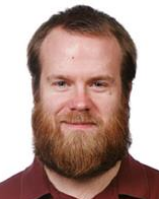}}]{Brian Hardy}~is a Principal Engineer in The Aerospace Corporation’s Small Satellite Department. Dr. Hardy joined The Aerospace Corporation in 2001 and has designed, performed, and analyzed data from a variety of fluid mechanics, structural mechanics, and thermal experiments, to develop new technologies as well as evaluate reliability and diagnose failures of existing spacecraft and launch systems. Dr. Hardy has been working for the past fifteen years to build, calibrate, test, and qualify sensors and subsystems for small satellites built by The Aerospace Corporation, and is presently the lead engineer in charge of integration, test, and on-orbit operations. He holds a B.S. in Mechanical Engineering from Caltech and a Ph.D. in Aerospace Engineering from UCLA.
\end{IEEEbiography}

\begin{IEEEbiography}[{\includegraphics[width=1in,height=1.25in,clip,keepaspectratio]{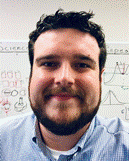}}]{Benjamen Bycroft}~is Director of Vehicle Autonomy \& System Trust at The Aerospace Corporation. His work spans research, development, and application of advanced and AI-enabled autonomy, dynamics and control, as well as electromechanical systems engineering for both launch vehicles and spacecraft. Benjamen has previously taught courses within the University of Southern California’s Viterbi School of Engineering. He holds a B.S. in Mechanical Engineering and an M.S. in Dynamics \& Control from USC, as well as a B.A. in Physics from Willamette University.
\end{IEEEbiography}

\begin{IEEEbiography}[{\includegraphics[width=1in,height=1.25in,clip,keepaspectratio]{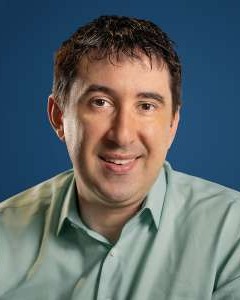}}]{Darren W. Rowen}~is Principal Engineer for Applied Edge Computing at The Aerospace Corporation's xLab, where he guides integration of edge computing with prototypes and spearheads the design of satellite and payload architectures to support autonomous operations. He earned his combined B.S./M.S. in Electrical Engineering from the Rochester Institute of Technology, focusing on control systems. Mr. Rowen joined Aerospace in 2005 as a Member of the Technical Staff specializing in control systems, and after working at Northrop Grumman Space Technology (2007-2011) supporting control systems analysis for Class A spacecraft, he rejoined Aerospace in 2011. He became the lead developer for one of the first CubeSat-size attitude determination and control systems and was instrumental in demonstrating the first space-to-ground optical communication link from a CubeSat. He served as Manager of the Small Satellite Development \& Operations section (2014) and Director of the Small Satellite Department (2018-2025), enabling the successful launch and operations of more than 25 custom small satellites forming one of the biggest privately managed satellite constellations.
\end{IEEEbiography}

\begin{IEEEbiography}[{\includegraphics[width=1in,height=1.25in,clip,keepaspectratio]{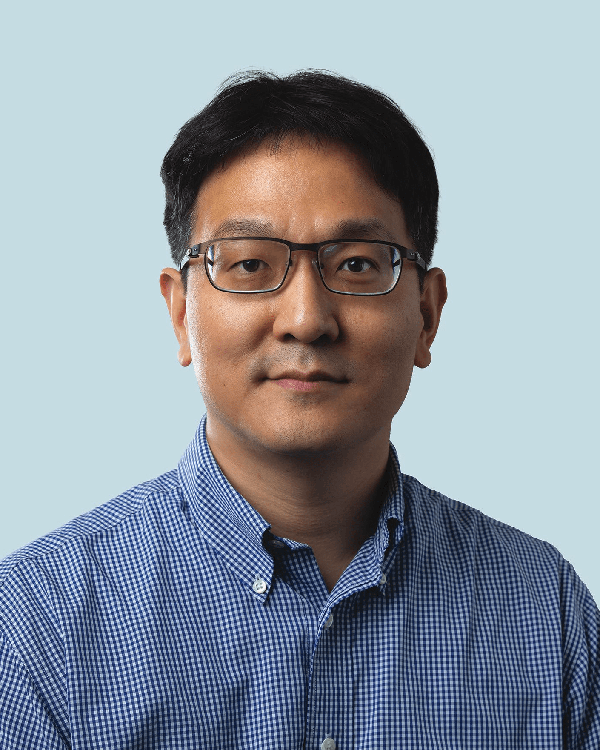}}]{Soon-Jo Chung}~(M’06–SM’12) received the S.M. degree in aeronautics and astronautics and the Sc.D. degree in estimation and control from Massachusetts Institute of Technology, Cambridge, MA, USA, in 2002 and 2007, respectively. He is currently Bren Professor of Control and Dynamical Systems, and a Jet Propulsion Laboratory Senior Research Scientist at the California Institute of Technology, Pasadena, CA, USA. 
Dr. Chung was the recipient of the UIUC Engineering Deans Award for Excellence in Research, the Beckman Faculty Fellowship of the UIUC Center for Advanced Study, the U.S. Air Force Office of Scientific Research Young Investigator Award, the National Science Foundation Faculty Early Career Development Award, a 2020 Honorable Mention for IEEE RA-L Best Paper Award, and three Best Conference Paper Awards from the IEEE and the American Institute of Aeronautics and Astronautics.
He was an Associate Editor of  IEEE Transactions on Automatic Control and IEEE Transactions on Robotics, and the Guest Editor of a Special Section on Aerial Swarm Robotics published in IEEE Transactions on Robotics.
\end{IEEEbiography}